\documentclass[sigconf]{aamas}

\usepackage{balance} 
\usepackage{subfig}
\usepackage{algorithm,algorithmic}

\usepackage{multirow}
\usepackage{makecell}

\makeatletter
\gdef\@copyrightpermission{
  \begin{minipage}{0.2\columnwidth}
   \href{https://creativecommons.org/licenses/by/4.0/}{\includegraphics[width=0.90\textwidth]{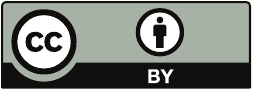}}
  \end{minipage}\hfill
  \begin{minipage}{0.8\columnwidth}
   \href{https://creativecommons.org/licenses/by/4.0/}{This work is licensed under a Creative Commons Attribution International 4.0 License.}
  \end{minipage}
  \vspace{5pt}
}
\makeatother

\setcopyright{ifaamas}
\acmConference[AAMAS '25]{Proc.\@ of the 24th International Conference
on Autonomous Agents and Multiagent Systems (AAMAS 2025)}{May 19 -- 23, 2025}
{Detroit, Michigan, USA}{Y.~Vorobeychik, S.~Das, A.~Nowé  (eds.)}
\copyrightyear{2025}
\acmYear{2025}
\acmDOI{}
\acmPrice{}
\acmISBN{}

\acmSubmissionID{<<OpenReview submission id>>}

\title{Offline-to-Online Multi-Agent Reinforcement Learning with Offline Value Function Memory and Sequential Exploration}

\author{Hai Zhong}
\affiliation{
  \institution{IIIS, Tsinghua University}
  \city{Beijing}
  \country{China}}
\email{zhongh22@mails.tsinghua.edu.cn}

\author{Xun Wang}
\affiliation{
  \institution{IIIS, Tsinghua University}
  \city{Beijing}
  \country{China}}
\email{wang-x24@mails.tsinghua.edu.cn}

\author{Zhuoran Li}
\affiliation{
  \institution{IIIS, Tsinghua University}
  \city{Beijing}
  \country{China}}
\email{lizr20@mails.tsinghua.edu.cn}

\author{Longbo Huang*} \thanks{*Corresponding to  Longbo Huang $\langle$\href{longbohuang@tsinghua.edu.cn}{longbohuang@tsinghua.edu.cn}$\rangle$. }
\affiliation{
  \institution{IIIS, Tsinghua University}
  \city{Beijing}
  \country{China}}
\email{longbohuang@tsinghua.edu.cn}

\begin{abstract}
Offline-to-Online Reinforcement Learning has emerged as a powerful paradigm, leveraging offline data for initialization and online fine-tuning to enhance both sample efficiency and performance. However, most existing research has focused on single-agent settings, with limited exploration of the multi-agent extension, i.e., Offline-to-Online Multi-Agent Reinforcement Learning (O2O MARL). In O2O MARL, two critical challenges become more prominent as the number of agents increases: (i) the risk of unlearning pre-trained Q-values due to distributional shifts during the transition from offline-to-online phases, and (ii) the difficulty of efficient exploration in the large joint state-action space. To tackle these challenges, we propose a novel O2O MARL framework called \textbf{Offline Value Function Memory with Sequential Exploration (OVMSE)}. First, we introduce the \textbf{Offline Value Function Memory (OVM)} mechanism to compute target Q-values, preserving knowledge gained during offline training, ensuring smoother transitions, and enabling efficient fine-tuning. Second, we propose a decentralized \textbf{Sequential Exploration (SE)} strategy tailored for O2O MARL, which effectively utilizes the pre-trained offline policy for exploration, thereby significantly reducing the joint state-action space to be explored. Extensive experiments on the StarCraft Multi-Agent Challenge (SMAC) demonstrate that OVMSE significantly outperforms existing baselines, achieving superior sample efficiency and overall performance.
\end{abstract}

\keywords{Multi-Agent Reinforcement Learning; Offline-to-Online Reinforcement Learning}

\newcommand{\BibTeX}{\rm B\kern-.05em{\sc i\kern-.025em b}\kern-.08em\TeX}

\begin{document}


\pagestyle{fancy}
\fancyhead{}


\maketitle 

\section{Introduction}

Multi-Agent Reinforcement Learning (MARL) has achieved remarkable success across various domains, including mastering complex video games \cite{grandmasterstarcraft2, Dota, EmergentToolUse}, optimizing warehouse logistics \cite{logisticswarehouse}, enabling robotic soccer \cite{deepmindscooer}, and performing bi-manual dexterous manipulation \cite{Bidexhands}. However, these successes often come at the cost of low sample efficiency and high computational overhead, as MARL algorithms must explore a joint state-action space that grows exponentially with the number of agents. A promising approach to alleviate this computational burden is Offline-to-Online (O2O) Reinforcement Learning (RL). In recent years, O2O RL has achieved significant progress \cite{Aperspectiveofqlearning,Unio4,PopulationO2O,BayesianO2O,Cal-QL,ENOTO,PEX,lee2021offlinetoonline}, leveraging advances in Offline RL by utilizing offline datasets \cite{d4rl,offthegrid} to provide strong initial policies and pre-trained value functions. By learning a high-quality pre-trained policy from offline data, single-agent O2O RL can significantly reduce the need for extensive exploration and further enhance performance through interaction during the online phase. However, existing results primarily focus on single-agent scenarios, and the important O2O MARL setting has received only very limited attention, e.g., \cite{MAO2O}.

\begin{figure}[htbp]
    \centering
\includegraphics[width=0.42\textwidth]{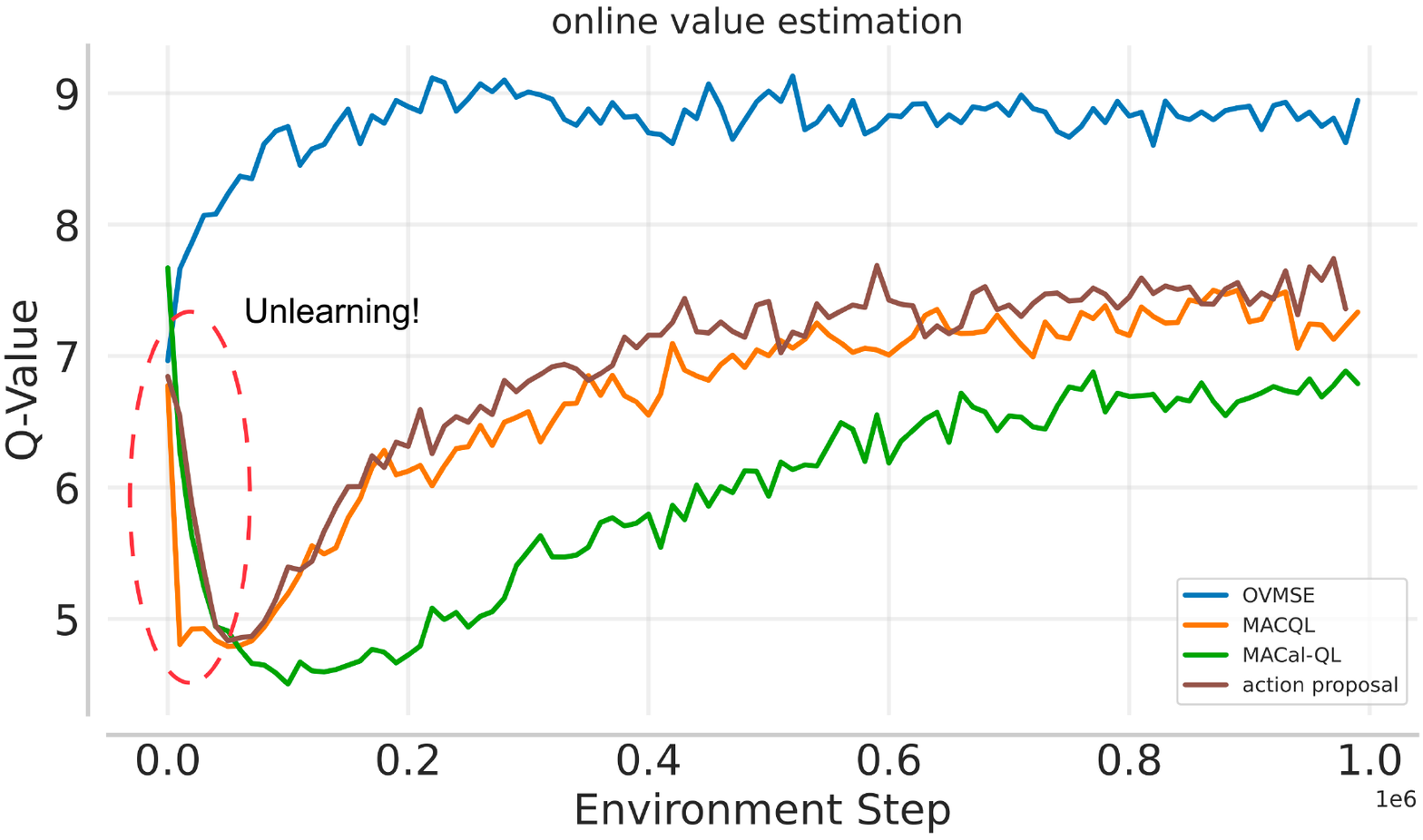}
    \caption{Q-value evolution for different algorithms during the online phase, starting from pre-trained Q-value from offline learning. 
    OVMSE clearly preserves the offline Q-value, while other methods experience rapid unlearning (i.e., online value estimations for these samples decrease rapidly).}

    \label{fig:1}
\end{figure}

O2O MARL faces two major challenges. First, exploration during the transition from the offline-to-online phase introduces a distributional shift, that can result in significant \emph{unlearning} of the pre-trained Q-values in the early stages of online learning. To demonstrate this phenomenon, we conducted an experiment analyzing how pre-trained Q-values evolve during online fine-tuning. We collected trajectories for the offline pre-trained agent interacting with the environment and stored them into a replay buffer. During the online phase, we evaluated how online value estimations for these samples evolved. As shown in Figure~\ref{fig:1}, Multi-agent Conservative Q-Learning (MACQL) \cite{CQL}, Multi-agent Cal-QL (MACal-QL) \cite{Cal-QL}, and the action proposal method from \cite{ImitationBoostrap} exhibit clear signs of unlearning, as the fine-tuned Q-values exhibit a rapid drop during the initial stage of online learning. This suggests that the algorithms have forgotten the previously learned optimal actions. This unlearning behavior hinders the efficiency of online fine-tuning, as the policy has to relearn knowledge that was already acquired during the offline phase. In contrast, our proposed algorithm OVMSE preserves the offline knowledge and achieves fast online fine-tuning.

Second, efficient exploration is critical for O2O MARL. Exploration at the initial stage of online learning can lead to overestimation of Q-values for unseen state-action pairs. While this overestimation is necessary for agents to experiment with new actions and improve upon the offline pre-trained policy, inefficient exploration may result in assigning high Q-values to sub-optimal actions, requiring significant trial and error to correct. This issue is particularly pronounced in multi-agent systems, where the joint state-action space grows exponentially with the number of agents. However, unlike training from scratch, O2O MARL benefits from a pre-trained offline policy, which provides a much stronger starting point than a randomly initialized policy. Therefore, an effective exploration strategy for O2O MARL needs to focus on exploring more efficiently within a reduced joint state-action space, rather than exhaustively searching through the entire space.

To address these two challenges, we propose a novel O2O MARL algorithm named \textbf{Offline Value Function Memory and Sequential Exploration (OVMSE)}. Specifically, OVMSE consists of two key components. The first component, \textbf{OVM}, introduces a novel way of computing target Q-values and solves the unlearning problem. With OVM, even when the offline pre-trained Q-values degrade due to distributional shifts during the early stages of online learning, the algorithm allows for the quick recovery of the offline pre-trained values. The second key component, \textbf{Sequential Exploration}, is designed to enable efficient exploration after transitioning to the online phase. Inspired by the sequential update mechanism \cite{MASEQROLLOut,OrderMatters}, SE restricts exploration to one agent at a time by allowing a single agent to select a random action while the others follow their respective policies. SE reduces the complexity of the joint state-action space and improves the quality of exploration, enabling more efficient fine-tuning and better performance. We further develop decentralized SE for decentralized execution. We empirically validate OVMSE on a collection of easy, hard, and super-hard tasks in SMAC. Our results demonstrate superior performance compared to baseline methods.

The paper is organized as follows. Section~\ref{related work} discusses related literature, Section~\ref{preliminary} introduces the background knowledge for MARL and QMIX \cite{QMIX}, Section~\ref{OVMSE section} describes the proposed OVMSE algorithm, and Section~\ref{experiment} presents the experimental results and discussions. The main contributions of our paper are summarized as follows:

\begin{enumerate}

    \item[(i)] \textbf{Key Challenges in O2O MARL:} We identify and analyze two key challenges in O2O MARL: (1) Offline-learned Q-values can be unlearned during the initial stage of online learning, leading to the degradation of offline pre-trained policies; (2) Inefficient exploration in the exponentially large joint state-action space can result in slow online fine-tuning.

    \item[(ii)] \textbf{OVMSE Algorithm:} We propose OVMSE, a novel framework that combines Offline Value Function Memory (OVM) and Sequential Exploration (SE) to address both the unlearning and exploration issues. OVMSE is designed to facilitate robust and efficient online fine-tuning, reducing the sample complexity while achieving superior performance.
    
    \item[(iii)] \textbf{Extensive Empirical Evaluation on SMAC:} We conduct extensive empirical evaluations of OVMSE on a range of tasks—spanning easy, hard, and super-hard difficulties—in the StarCraft Multi-Agent Challenge (SMAC). Our results demonstrate that OVMSE significantly outperforms baseline methods. OVMSE shows significantly improved online performance, higher sample efficiency, and faster fine-tuning, underscoring its practical applicability to complex multi-agent environments.
    
\end{enumerate}


\section{RELATED WORK} \label{related work}

\textbf{O2O RL and O2O MARL.} Many previous works on O2O RL focus on reusing offline data during online training \cite{RLPD, lee2021offlinetoonline, BayesianO2O}. \cite{PEX} proposes retaining a copy of the offline policy and using a hybrid of the offline and online fine-tuned policies during online training, thereby preserving the performance achieved during the offline phase. A related approach to our work is presented in \cite{ImitationBoostrap} which utilizes the offline policy to suggest actions for online target value estimation. In the context of O2O MARL, relatively little research attention has been given. \cite{MAO2O} investigates O2O MARL by using offline data during the online phase and proposes metrics for sampling from offline data. In addition to our proposed value function memory, we employ a multi-agent sequential exploration strategy inspired by multi-agent sequential rollout \cite{MASEQROLLOut,OrderMatters} to reduce the exploration space in cooperative multi-agent systems. This improves online learning efficiency, as the offline pre-trained policy can focus on targeted exploration rather than an exhaustive random search of the action space, which is typically required when training from scratch.

\textbf{Offline MARL.} The principles used to address accumulated extrapolation error in offline single-agent RL also apply to offline MARL. A common strategy is to incorporate pessimism towards out-of-distribution (OOD) states and actions, which has proven effective in the offline MARL setting as well. Both \cite{OMAR} and \cite{CFCQL} extend the Conservative Q-Learning (CQL) framework \cite{CQL} to the multi-agent domain. \cite{OMAR} identifies that value functions in MARL are more prone to local optima as the number of agents increases and proposes using zeroth-order optimization to avoid being trapped in these local optima. \cite{CFCQL} demonstrates that independently penalizing each agent's OOD actions can reduce the overall conservatism in value estimation, leading to improved empirical performance. \cite{MAICQ} takes a different approach by ensuring that only actions present in the dataset are used for target Q-value estimation, thereby promoting conservatism for OOD states and actions. Another promising direction is to treat offline MARL problems as supervised learning tasks. For example, \cite{MAKD} and \cite{MADT} frame offline MARL as sequence modeling problems, leveraging the power of decision transformers \cite{DecisionTransformer} to achieve strong empirical performance. Furthermore, \cite{BeyondConservatism} and \cite{MADiff} employ diffusion models as policy backbones, demonstrating good performance and the ability to learn diverse strategies. 


\section{PRELIMINARIES} \label{preliminary}

\subsection{Multi-Agent Reinforcement Learning Formulation}
A cooperative MARL problem can be described as a decentralized partially observable Markov decision process (Dec-POMDP) \cite{IntroDecPomdp}, which can be represented as a tuple \( \langle \mathcal{N}, S, O, A, P, r, \gamma, T \rangle \).

\begin{itemize}
    \item \( \mathcal{N} = \{1, 2, \ldots, N\} \) is the set of agents.
    \item  S  is the set of states of the environment.
    \item \( O = O_1 \times O_2 \times \ldots \times O_N \) is the joint observation space, where \( O_i \) is the set of observations available to agent \( i \).
    \item \( A = A_1 \times A_2 \times \ldots \times A_N \) is the joint action space, where \( A_i \) is the set of actions available to agent \( i \).
    \item Given the current state \( \boldsymbol{s} \in S \) and the joint action \( \boldsymbol{a} \in A \), \( P(\boldsymbol{s}' \mid \boldsymbol{s}, \boldsymbol{a}) \) and \( r(\boldsymbol{s}, \boldsymbol{a}) \) describe the probability of transitioning from \( \boldsymbol{s} \) to the next state \( \boldsymbol{s}' \), and the immediate reward shared by all agents, respectively.
    \item \( \gamma \in [0, 1] \) is the discount factor, and \( T \) is the time horizon of the process.
\end{itemize}

At time step \( t \), each agent maintains its observation-action history \( \tau_i \in \mathcal{T}_i = \{(o_{i,0}, a_{i,0}, \ldots, o_{i,t}, a_{i,t})\} \), and the joint observation-action history is \( \tau = (\tau_1, \tau_2, \ldots, \tau_n) \). The objective is to find a policy for each agent \( \pi_i: \mathcal{T}_i \to A_i \) that maximizes the expected discounted return
\[
\mathbb{E} \left[ \sum_{t=0}^{T} \gamma^t \cdot r(\boldsymbol{s}_t, \boldsymbol{a}_t) \right],
\]
where \( \boldsymbol{s}_t \) and \( \boldsymbol{a}_t \) are the state and the joint action at time step \( t \), respectively.


\subsection{QMIX} 
In this section, we present QMIX \cite{QMIX,QMIXJMLR}, which is a popular MARL algorithm following the Centralized Training with Decentralized Execution (CTDE) paradigm. In our work, we utilize QMIX as the backbone for both the offline and online phases.

Under the CTDE framework, each agent has an individual action-value function \( Q_i({\tau}_i, a_i) \), based on its observation-action history. Additionally, QMIX introduces a joint action-value function \( Q_{\text{tot}}({\tau}, \boldsymbol{a}) \), constructed using a mixing network \( f_s \), which combines individual agent action-values according to:
\begin{equation}
Q_{\text{tot}}({\tau}, \boldsymbol{a}) = f_s(Q_1({\tau}_1, a_1), \ldots, Q_n({\tau}_n, a_n)).
\end{equation}

The mixing network satisfies the monotonicity constraint,
\begin{equation}
\frac{\partial Q_{\text{tot}}}{\partial Q_i} \geq 0 \quad \forall i \in \mathcal{N},
\end{equation}
thereby maintaining that the joint action-value function \( Q_{\text{tot}} \) is a non-decreasing function with respect to each individual action-value function \( Q_i \). Hence, QMIX satisfies the Individual-Global-Maximum (IGM) property \cite{Qtran}, which guarantees that the joint optimal action can be constructed by combining each agent's individual optimal actions based on their local observations:
\begin{equation}
\arg\max_{\boldsymbol{a}} Q_{\text{tot}}(\tau, \boldsymbol{a}) = \left( \arg\max_{a_1} Q_1(\tau_1, a_1), \ldots, \arg\max_{a_n} Q_n(\tau_n, a_n) \right).
\end{equation}

\section{ Offline Value Function Memory and Sequential Exploration (OVMSE)} \label{OVMSE section}
We now introduce our Offline Value Function Memory and Sequential Exploration (OVMSE) algorithm for O2O MARL. First, we present our online training mechanism, OVM. Next, we describe the designed decentralized sequential exploration mechanism, SE. Finally, we outline our offline training procedure.


\subsection{Offline Value Function Memory} 

\label{OVM}
In the online phase, we introduce our Offline Value Function Memory (OVM) target. We describe the functionality of our proposed OVM training objective for online training. After offline training, we retain a copy of the pre-trained target value function, denoted as $\bar{Q}_{\text{tot-offline}}$. We introduce the OVM target, defined as:
\begin{equation} 
    \bar{Q}_{\text{OVM}} = \max \left( \bar{Q}_{\text{tot-offline}}(\tau, \boldsymbol{a}),\ r + \gamma \max_{\boldsymbol{a}'} \bar{Q}_{\text{tot}}(\tau', \boldsymbol{a}') \right).
\end{equation}
where $\bar{Q}_{\text{tot}}$ is the online target Q-function, $\tau'$ and $\boldsymbol{a}'$ are the joint observation-action history and joint action at the next time step, $r + \gamma \max_{\boldsymbol{a}'} \bar{Q}_{\text{tot}}(\tau', \boldsymbol{a}')$ is the online temporal difference target, and $\bar{Q}_{\text{tot-offline}}(\tau, \boldsymbol{a})$ represents the offline memory of the agents with joint history $\tau$ and joint action $\boldsymbol{a}$. The OVM target selects the maximum value between the offline value function memory and the online temporal difference target.

During the online phase, the value function is trained to minimize the mean squared error (MSE) between its estimation and the OVM target, as well as the MSE between its estimation and the online temporal difference target, i.e.,
\begin{equation}\label{loss_ovm}
\begin{aligned}
\mathcal{L}_{\text{OVM}} &= (1 - \lambda_{\text{memory}}) \left( Q_{\text{tot}}(\tau, \boldsymbol{a}) 
- \left( r + \gamma \max_{\boldsymbol{a}'} \bar{Q}_{\text{tot}}(\tau', \boldsymbol{a}') \right) \right)^2 \\
&\quad + \lambda_{\text{memory}} \left( Q_{\text{tot}}(\tau, \boldsymbol{a})  
- \bar{Q}_{\text{OVM}} \right)^2,
\end{aligned}
\end{equation}
which introduces a trade-off between the OVM target and the online temporal difference target. Here, $\lambda_{\text{memory}}$ is a memory coefficient that controls the balance between these two objectives.

We further propose an annealing schedule for $\lambda_{\text{memory}}$, which gradually adjusts the influence of the offline value memory over time. The annealing schedule is defined as follows: \footnote{We clarify that the annealing schedules in Eq. (6) and Eq. (7) utilize the $\max$ operator to ensure the parameters are lower-bounded by their terminal values, correcting a typo in an earlier version that used $\min$. We also rewrite the annealing schedule in an equivalent form to perfectly match implementation.}
\begin{align}\label{lambda_update}
\lambda_{\text{memory}} &= \max \Bigg( \lambda_{\text{memory\_end}}, \, \left( 1 - \left( \frac{1}{T} \right) \cdot t \right) \Bigg)
\end{align}
where $\lambda_{\text{memory\_end}}$ specifies the final value for $\lambda_{\text{memory}}$, $T$ and $\lambda_{\text{memory\_end}}$ determine the annealing duration, and $t$ represents the current environment step. The motivation for gradually decreasing $\lambda_{\text{memory}}$ is to maintain the memory of the offline value memory during the transition from offline-to-online learning. As the agent interacts with the environment and stores experiences in the replay buffer, the offline values are progressively adjusted towards the online temporal difference target values, facilitating further improvement.

This training objective ensures that the online value function preserves the pre-trained offline values while allowing the agent to follow the online temporal difference target when it is larger. As a result, the agent retains knowledge from the offline phase while effectively exploring new strategies and improving its performance during the online phase. The intuition behind OVM is as follows: the Q-values evolve based on the pre-trained offline Q-values. For unseen actions that were assigned low Q-values during offline training, their values can increase as they are explored online, allowing the agent to discover and exploit better strategies while also retaining offline knowledge.
\begin{figure*}[htbp]
    \centering
    \subfloat[]{\includegraphics[width=0.33\textwidth]{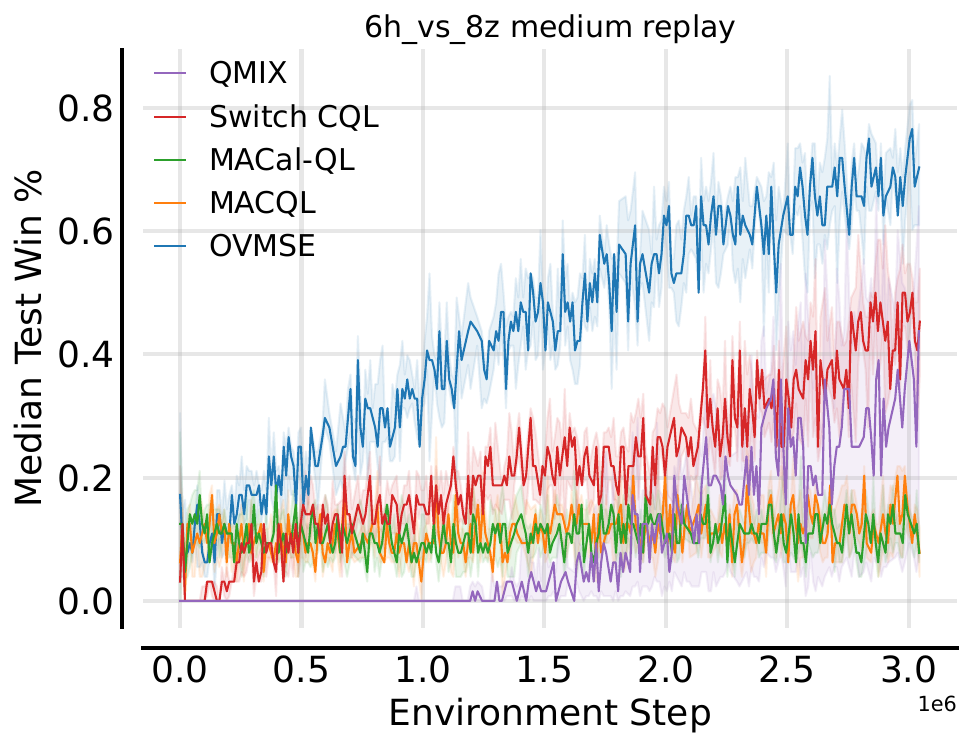}} 
    \subfloat[]{\includegraphics[width=0.33\textwidth]{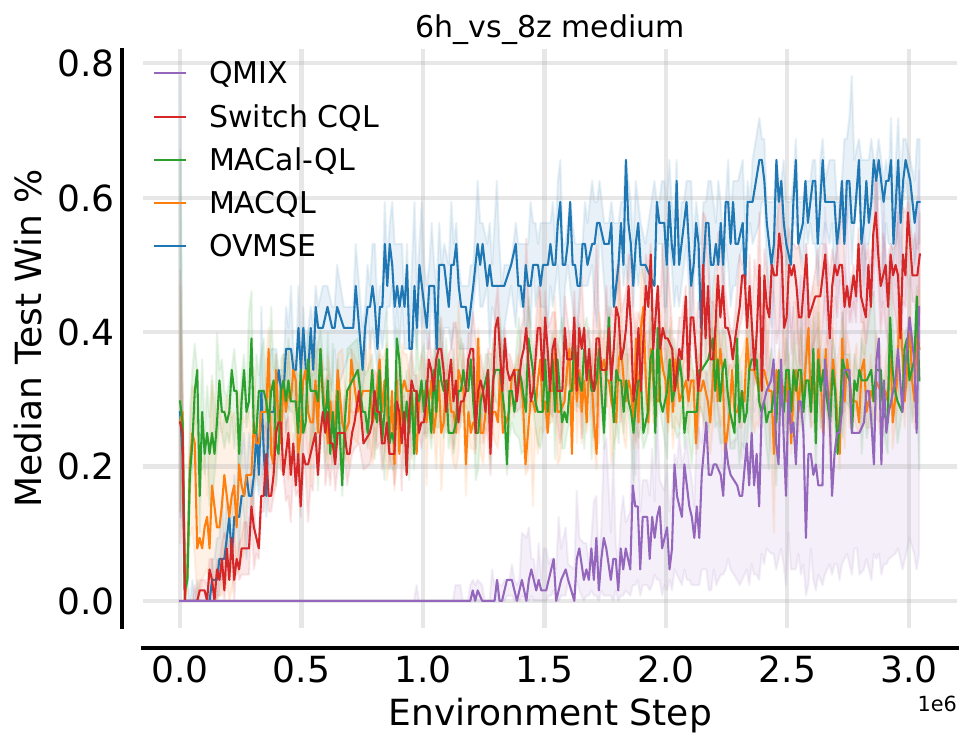}} 
    \subfloat[]{\includegraphics[width=0.33\textwidth]{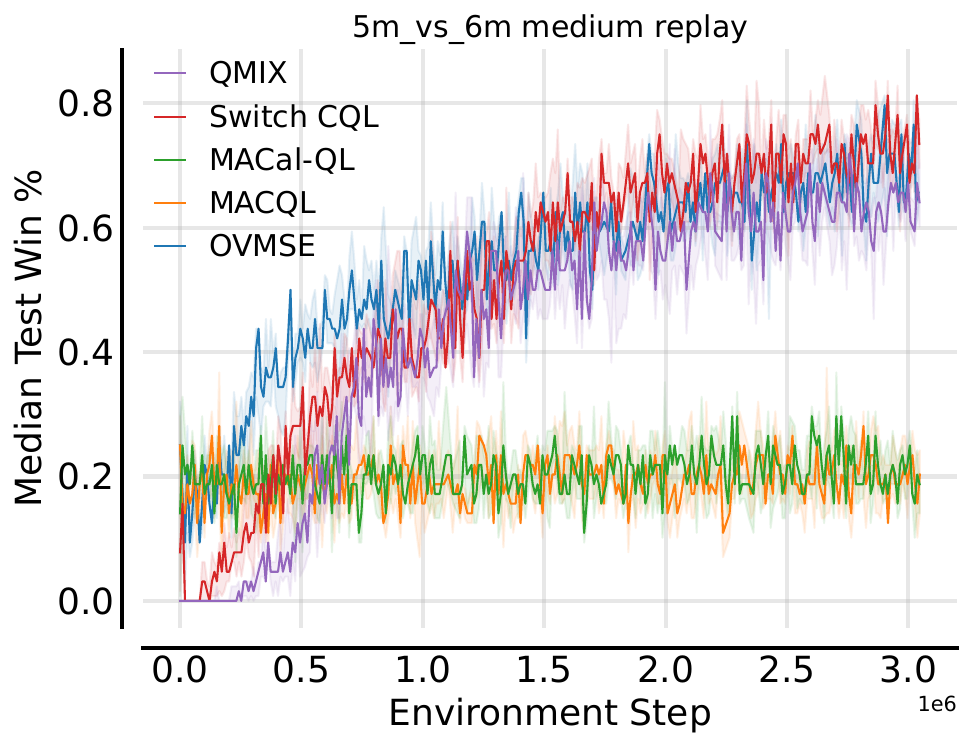}}
    \quad
    \subfloat[]{\includegraphics[width=0.33\textwidth]{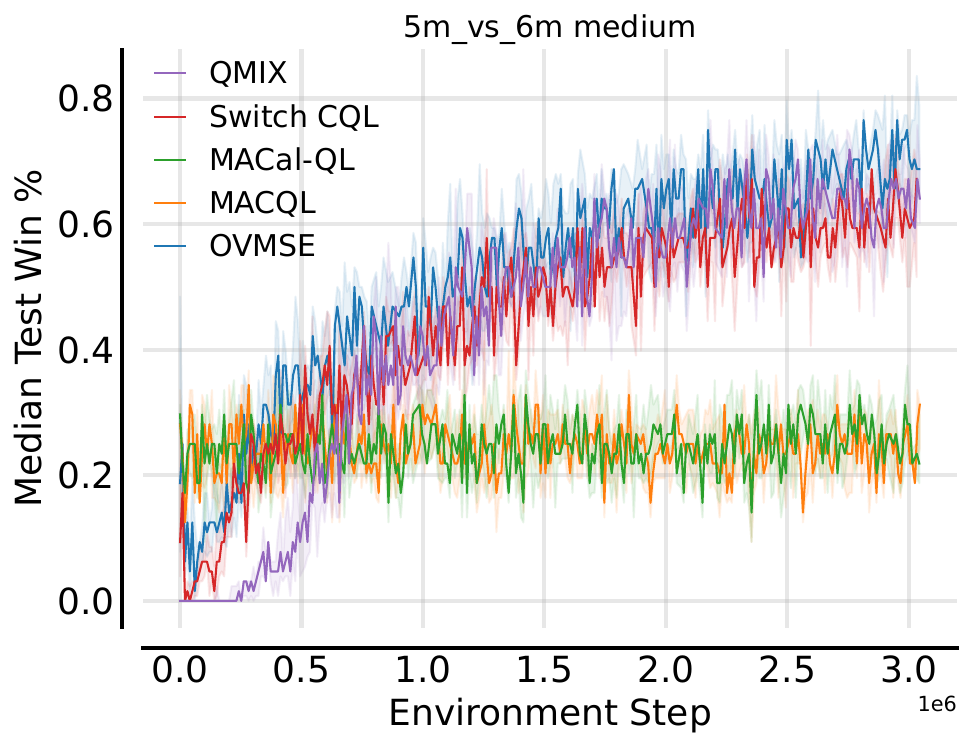}}
    \subfloat[]{\includegraphics[width=0.33\textwidth]{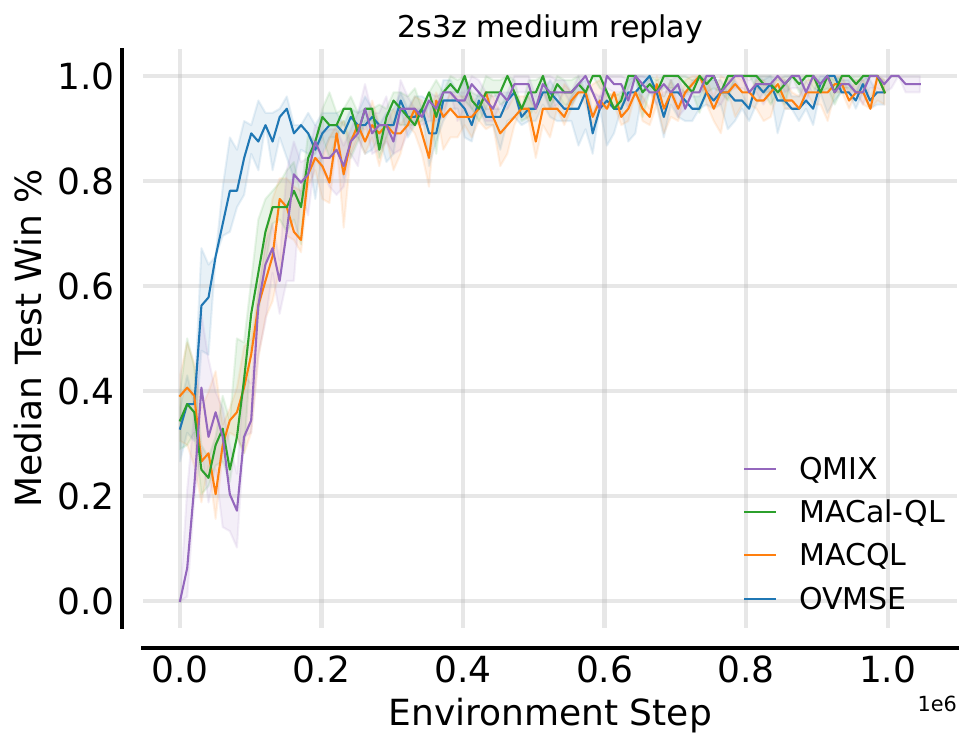}} 
    \subfloat[]{\includegraphics[width=0.33\textwidth]{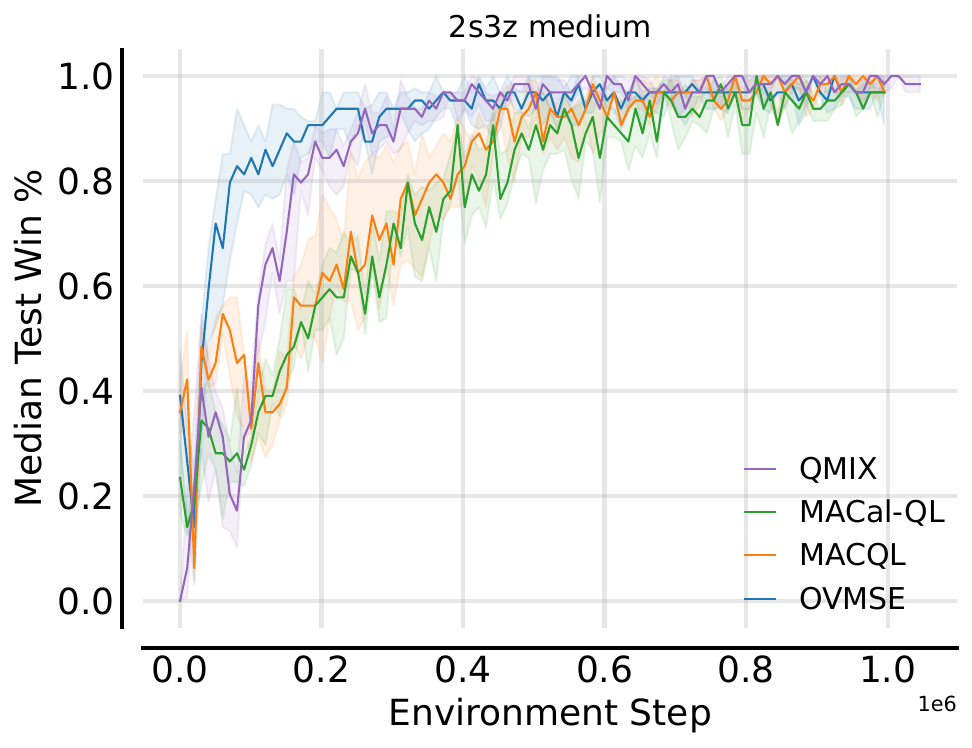}} 
    \quad
    \subfloat[]{\includegraphics[width=0.33\textwidth]{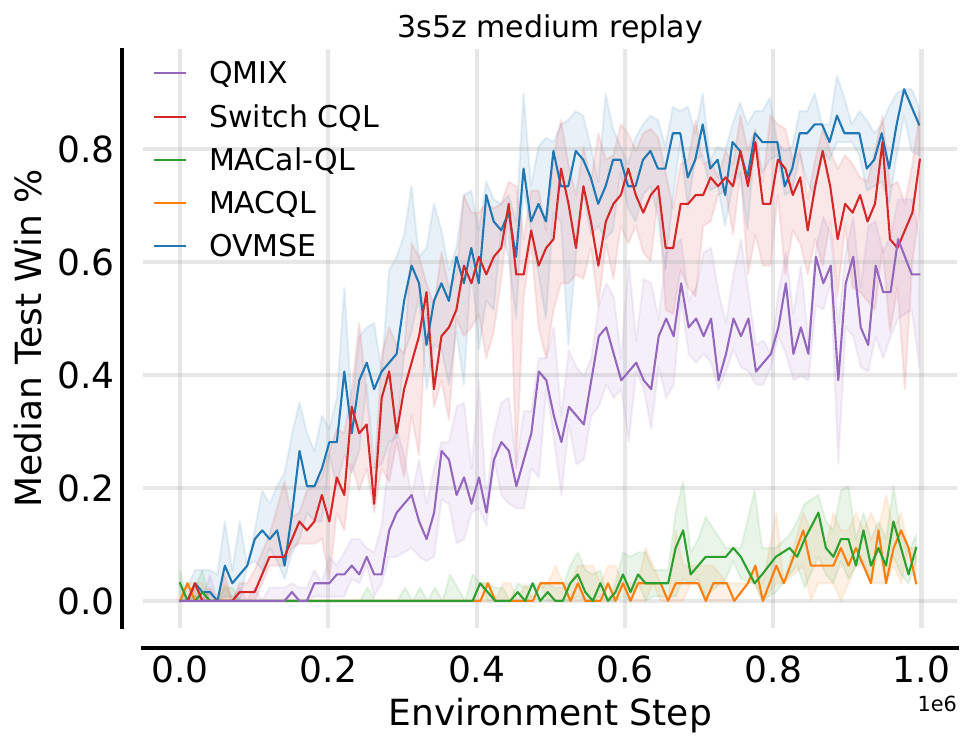}}
    \subfloat[]{\includegraphics[width=0.33\textwidth]{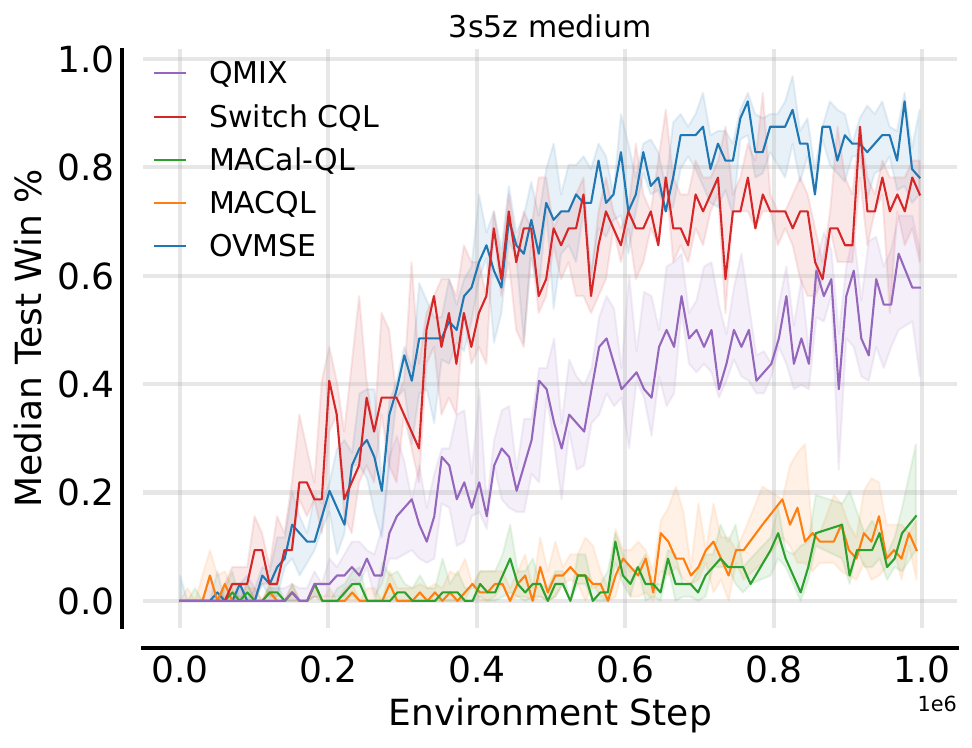}}
\caption{Median test win rates with respect to environment steps for OVMSE and baselines across different tasks, pre-trained with medium or medium replay datasets. Our results demonstrate that OVMSE preserves offline knowledge, enables efficient exploration, and leads to faster fine-tuning and superior overall performance.}

    \label{result figure}
\end{figure*}

\subsection{Multi-agent Sequential Exploration} \label{SE}

$\epsilon$-greedy exploration is one of the most widely used strategies for exploration in multi-agent reinforcement learning (MARL) algorithms \cite{hu2023rethinkingtheimplementation, QMIX}. In this strategy, each agent independently selects a random action with probability $\epsilon_t$, which gradually decreases over environment steps:
\begin{equation}\label{eps_update}
\epsilon_t = \max \Bigg( \epsilon_{\text{end}}, \, \left( 1 - \left( \frac{\epsilon_{\text{start}} - \epsilon_{\text{end}}}{T} \right) \cdot t \right) \Bigg),
\end{equation}
where $\epsilon_{\text{start}}$ and $\epsilon_{\text{end}}$ specify the start and terminal values for $\epsilon_t$, $T$ is the annealing duration, and $t$ is the current environment step. During the initial online phase, when $\epsilon$ is high, multiple agents are likely to explore randomly and simultaneously, resulting in exploration that resembles a random search across the exponentially large joint state-action space. However, this approach is inefficient for O2O MARL, where a pre-trained offline policy already provides a strong starting point. Instead of exhaustively searching the entire joint state-action space, exploration should focus on a smaller, more targeted subset to refine the policy rather than starting from scratch.

Inspired by the multi-agent sequential update scheme \cite{MASEQROLLOut, OrderMatters}, which sequentially updates different agents' policies, we introduce the concept of sequential exploration for O2O MARL. Specifically, we propose a sequential $\epsilon$-greedy exploration strategy. At each time step, agents collectively decide whether to explore based on the probability $\epsilon_t$, which decays with the number of environment steps. If exploration is chosen, only one randomly selected agent performs a random action, while all other agents act greedily, following the current policy. This coordinated approach allows the system to efficiently leverage the pre-trained policy while facilitating targeted exploration to further refine and improve the policy during the online phase.

\textbf{Decentralized Sequential Exploration.} However, this coordinated exploration requires communication between agents during execution, which may not always be feasible in decentralized execution settings. To address this limitation, we introduce a decentralized version of sequential exploration with a simple modification. We define a decentralized exploration probability, $\epsilon_{\text{dec\_t}} = \epsilon_t / N$, where $N$ is the number of agents. Each agent independently decides to explore with probability $\epsilon_{\text{dec\_t}}$, ensuring that, on average, the number of agents exploring at any given time matches that of the centralized version. This decentralized approach maintains the benefits of sequential exploration while ensuring compatibility with decentralized execution.

\begin{algorithm}[t]
\caption{Offline Value Function Memory with Sequential Exploration (OVMSE)}
\label{alg:full_ovmse}
\begin{algorithmic}[1]

\STATE Offline pretrain via a CQL-based algorithm and retain $\bar{Q}_{\text{OVM}}$.
\STATE \textbf{Initialize:} $\lambda_{\text{memory}} \leftarrow 1.0$, $\epsilon \leftarrow \epsilon_{\text{start}}$
\FOR{each environment step $t$}
    \STATE Explore the environment via SE (Algorithm \ref{alg:decentralized_sequential_exploration}).
    \STATE Draw samples from the offline dataset and online buffer.
    \STATE Calculate $\mathcal{L}_{\text{OVM}}$ according to \eqref{loss_ovm}.
    \STATE Update agents via QMIX. 
    \STATE Update $\lambda_{\text{memory}}$ and $\epsilon_t$ according to \eqref{lambda_update} and \eqref{eps_update}.
\ENDFOR
\end{algorithmic}
\end{algorithm}

\begin{algorithm}[t]
\caption{Decentralized Multi-Agent Sequential $\epsilon$-Greedy Exploration}
\label{alg:decentralized_sequential_exploration}
\begin{algorithmic}[1]
\STATE \textbf{Input:} exploration rate $\epsilon_t$ at environment step $t$
\STATE Set decentralized exploration probability: $\epsilon_{\text{dec\_t}} = \epsilon_t / N$
\FOR{each environment step $t$}
    \FOR{each agent $i \in \{1, 2, \dots, N\}$}
        \STATE \textbf{(Agents operate in parallel)}
        \STATE Sample $u \sim \text{Uniform}(0,1)$
        \IF{$u < \epsilon_{\text{dec\_t}}$}
            \STATE Agent $i$ selects a random action 
        \ELSE
            \STATE Agent $i$ selects a greedy action
        \ENDIF
    \ENDFOR
\ENDFOR
\end{algorithmic}
\end{algorithm}

\subsection{Offline Training for OVMSE}\label{sec:ovmse-offline}

In this section, we introduce the offline training procedure for OVMSE. During offline training, we utilize QMIX and Conservative Q-Learning (CQL) as our backbone algorithms. First, we minimize the squared temporal difference error as part of our offline training objective:
\begin{equation}
\mathcal{L}_{\text{QMIX}}=\left[ Q_{\text{tot}}({\tau}, \boldsymbol{a}) - \left( r + \gamma \max_{\boldsymbol{a}'} \bar{Q}_{\text{tot}}({\tau}', \boldsymbol{a}') \right) \right]^2, \label{QMIX_Training_Objective}
\end{equation}
where \( {\tau}' \) and \( \boldsymbol{a}' \) denote the next observation-action history and joint action, respectively, and $\bar{Q}_{\text{tot}}$ represents the target joint action-value function.

In addition to minimizing the temporal difference error, we incorporate the CQL training objective into the offline phase. CQL-based methods introduce a regularization term to the Q-value function's training objective, defined as:
\begin{equation}
\mathcal{L}_{\text{CQL}}=\mathbb{E}_{\tau \sim D, \boldsymbol{a} \sim \mu} \left[ Q(\tau, \boldsymbol{a}) \right] - \mathbb{E}_{\tau \sim D, \boldsymbol{a} \sim D} \left[ Q(\tau, \boldsymbol{a}) \right], \label{CQL_Regularization}
\end{equation}
where $\mu$ denotes the sampling distribution (e.g., sampling from the current policy or a uniform distribution). This regularization term intuitively penalizes the Q-values of actions not present in the dataset while promoting the Q-values of actions within the dataset.

The overall offline training objective combines the QMIX temporal difference error and the CQL regularization:
\begin{equation}\label{Offline_Training_Objective}
\mathcal{L}_{\text{offline}} = \mathcal{L}_{\text{QMIX}} + \alpha \mathcal{L}_{\text{CQL}},
\end{equation}
where $\alpha$ is a hyperparameter that controls the strength of the CQL regularization.

\begin{figure*}[htbp]
    \centering
    \subfloat[]{\includegraphics[width=0.33\textwidth]{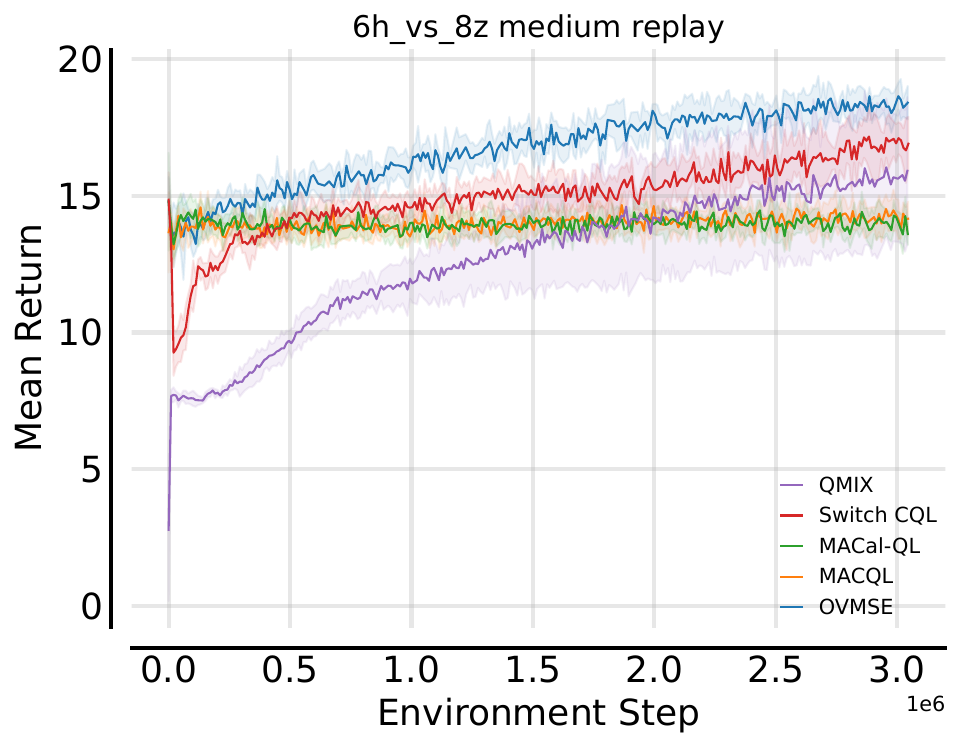}} 
    \subfloat[]{\includegraphics[width=0.33\textwidth]{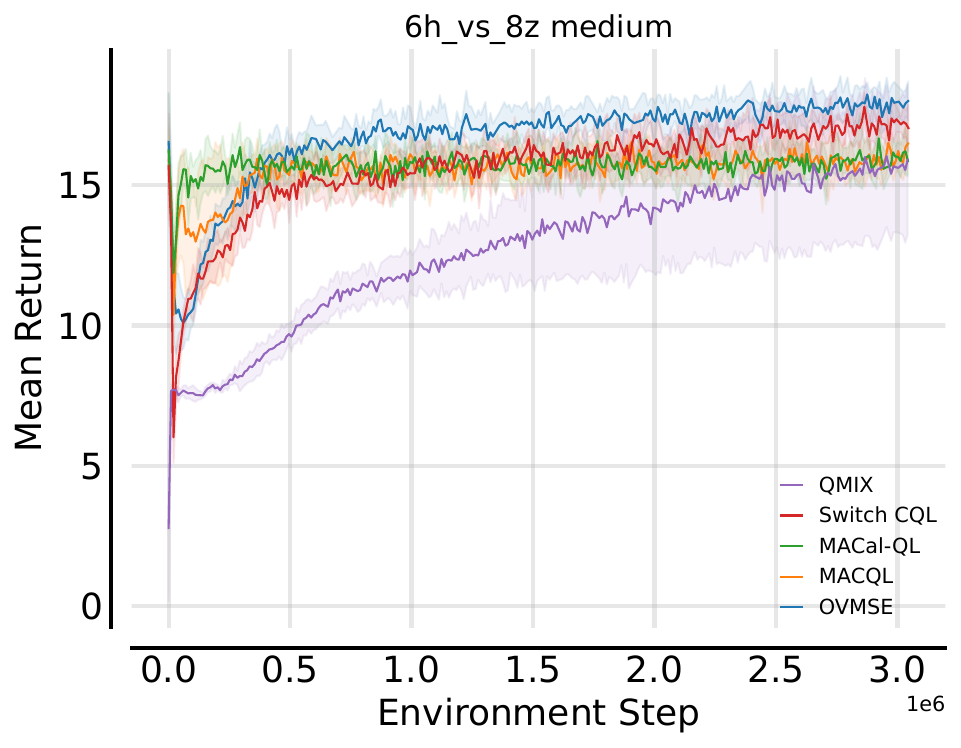}} 
    \subfloat[]{\includegraphics[width=0.33\textwidth]{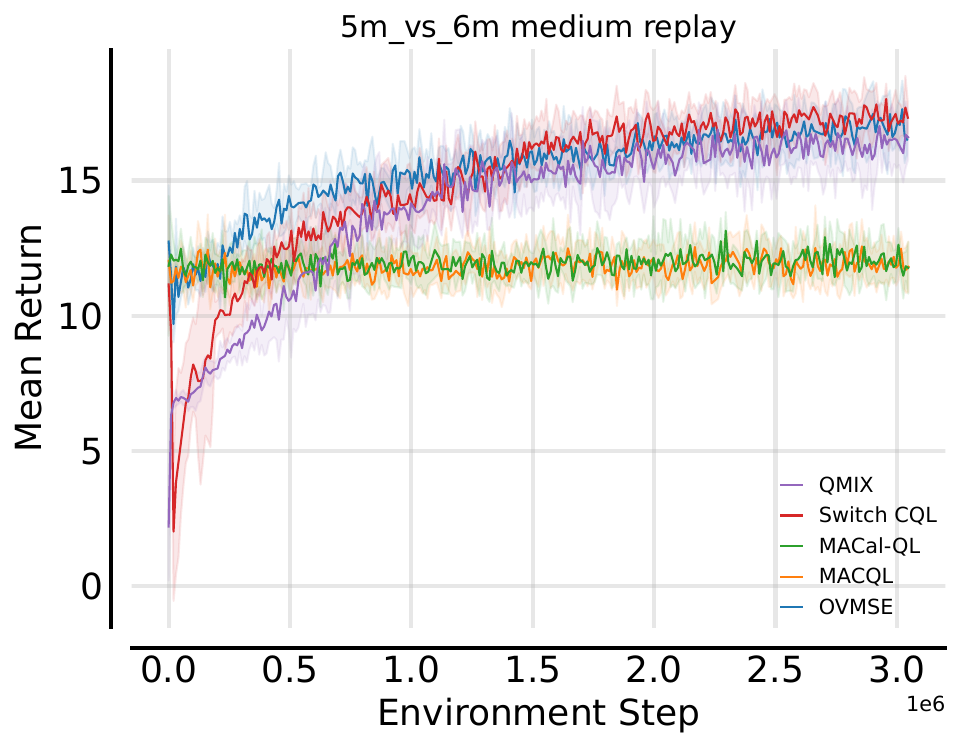}}
    \quad
    \subfloat[]{\includegraphics[width=0.33\textwidth]{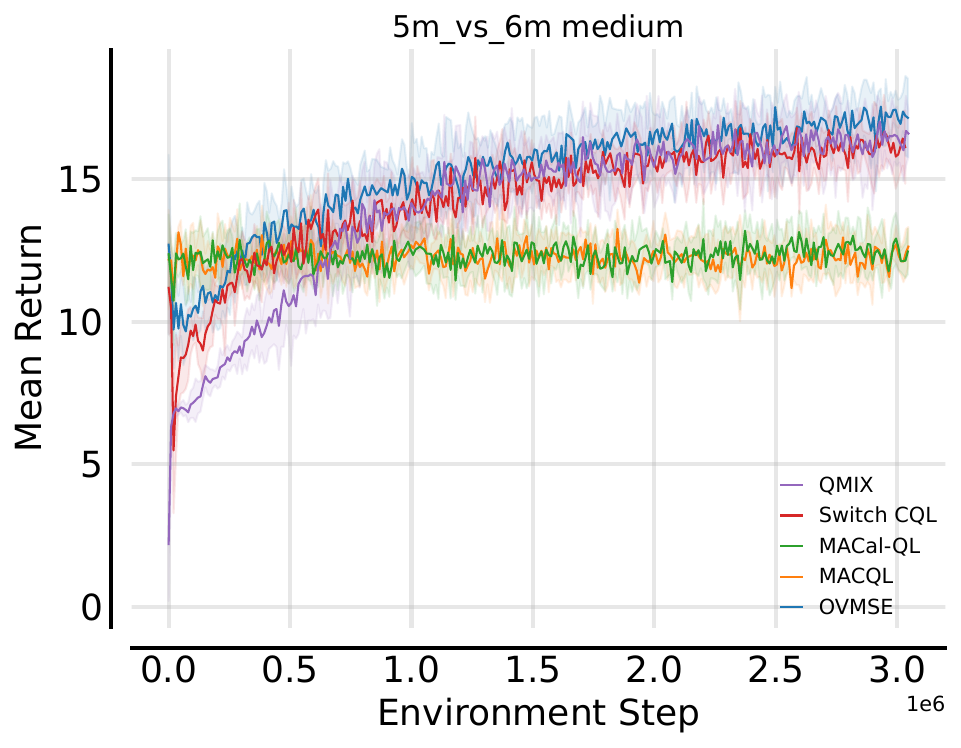}}
    \subfloat[]{\includegraphics[width=0.33\textwidth]{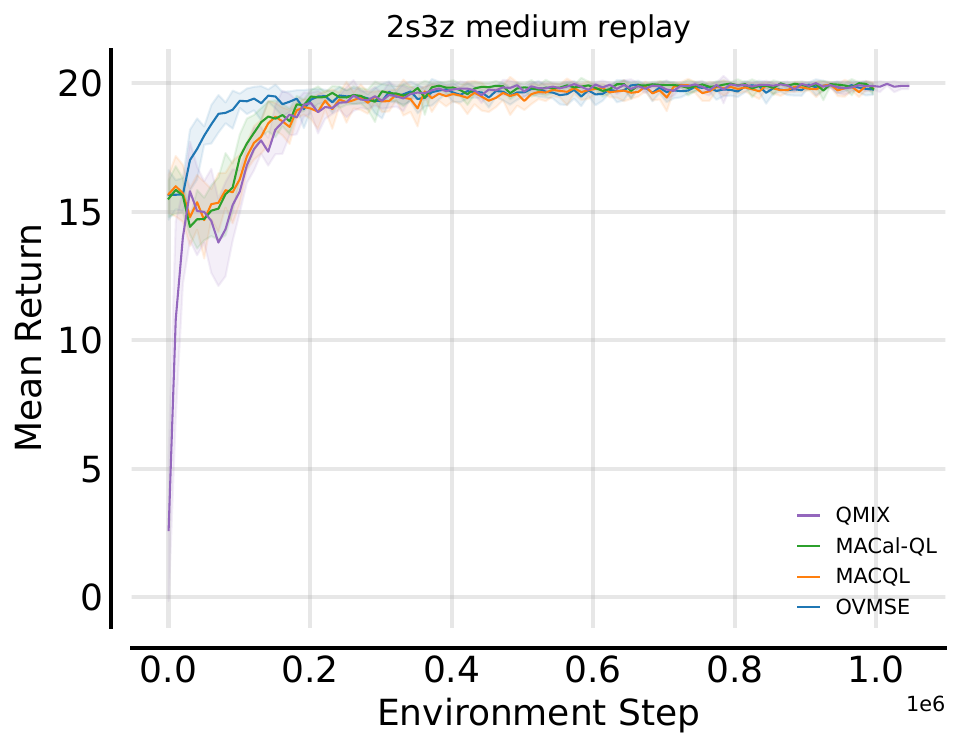}} 
    \subfloat[]{\includegraphics[width=0.33\textwidth]{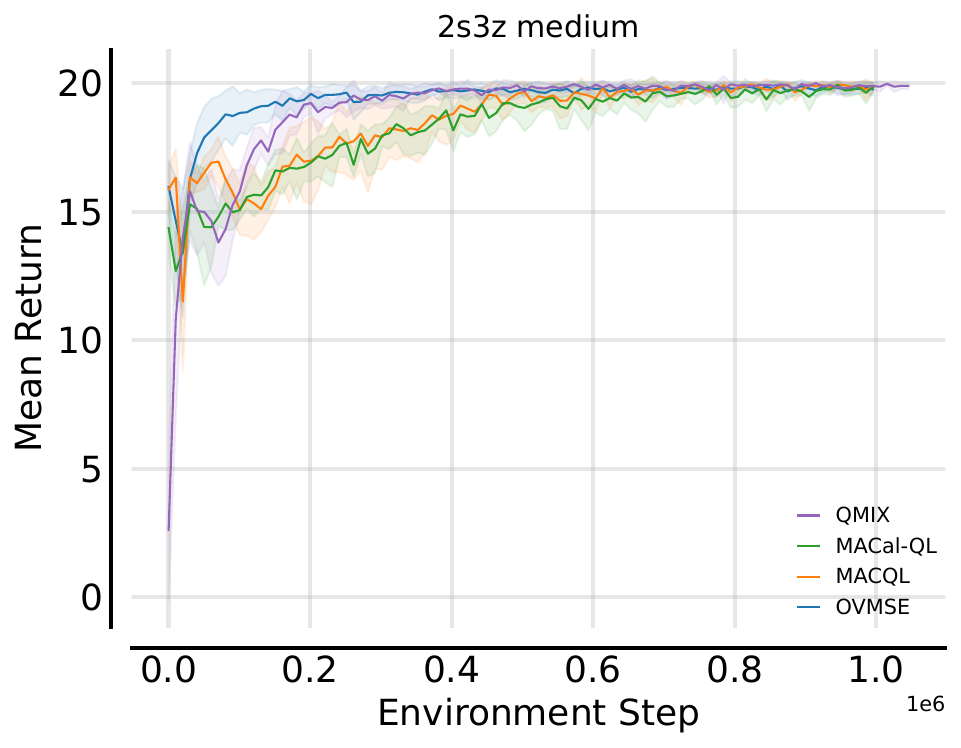}} 
    \quad
    \subfloat[]{\includegraphics[width=0.33\textwidth]{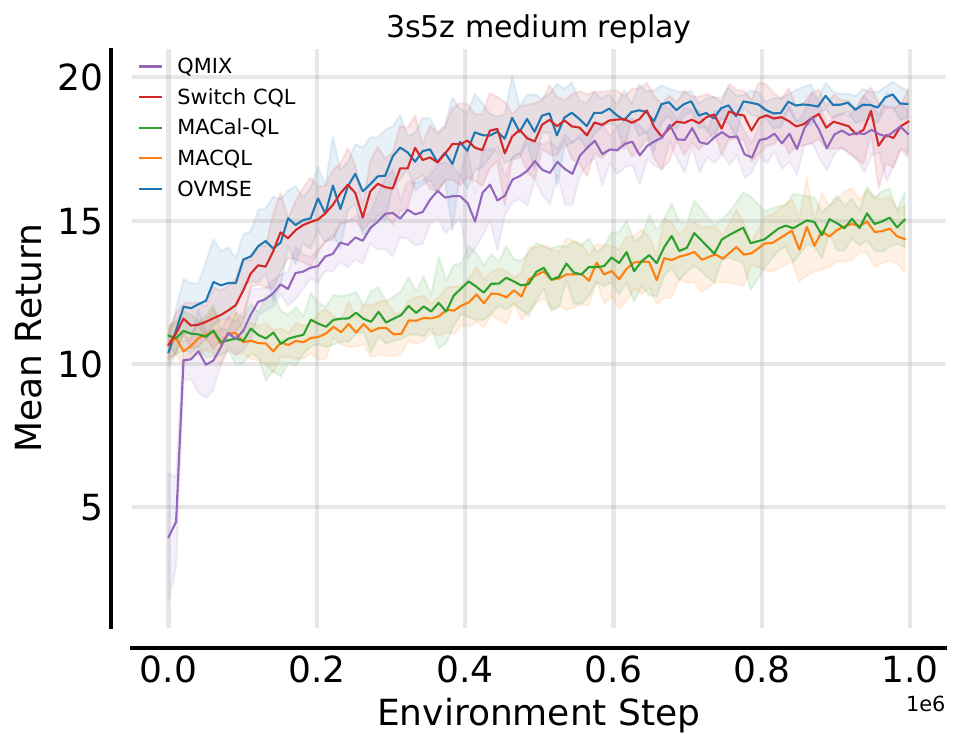}}
    \subfloat[]{\includegraphics[width=0.33\textwidth]{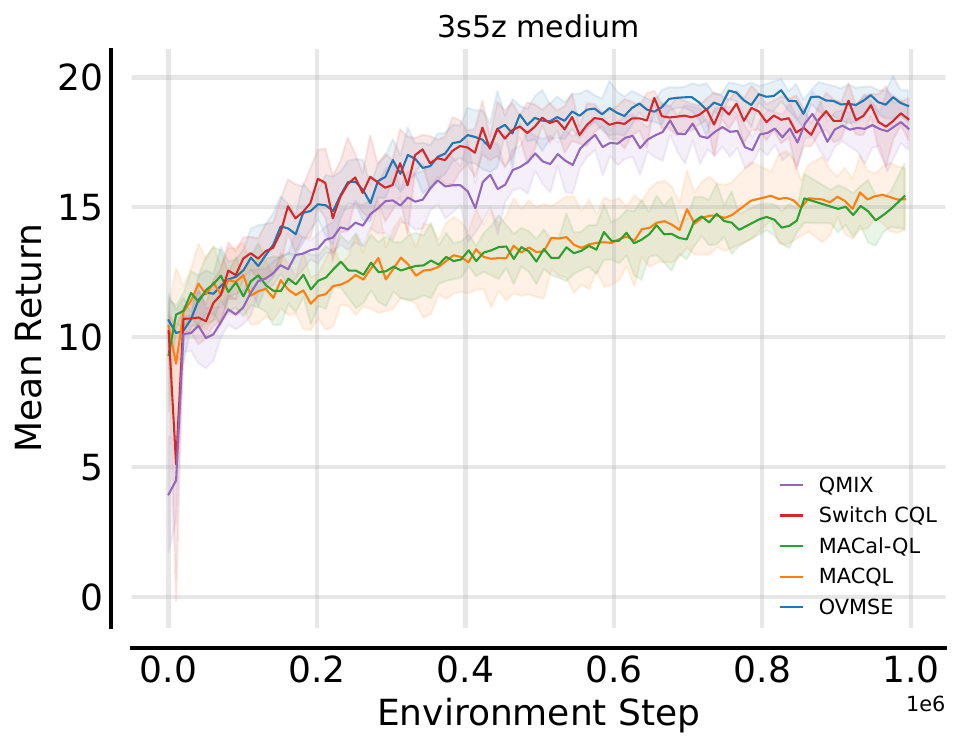}}
\caption{Mean test returns for all tasks under different algorithms.}

    \label{test return}
\end{figure*}

\section{EXPERIMENT} \label{experiment}
\subsection{Setup}

\textbf{Codebase, Offline Datasets, and Experiment Setup.} We evaluate our proposed OVMSE algorithm using the StarCraft Multi-Agent Challenge (SMAC), a key benchmark for MARL algorithms, across four tasks: \textit{2s3z} (easy), \textit{3s5z} (easy), \textit{5m\_vs\_6m} (hard), and \textit{6h\_vs\_8z} (super hard). For the tasks \textit{2s3z}, \textit{5m\_vs\_6m}, and \textit{6h\_vs\_8z}, we use offline datasets provided by CFCQL \cite{CFCQL}. We collected our own dataset for \textit{3s5z}, as it is not publicly available. Offline training leverages medium and medium-replay datasets \cite{d4rl}.  During the online phase, we apply the CQL loss for OVMSE in \textit{2s3z}, as it stabilizes online training. However, for \textit{3s5z}, \textit{5m\_vs\_6m} and \textit{6h\_vs\_8z}, we exclude the CQL loss because it hinders online learning by preventing the discovery of new strategies. Our implementation, including code and hyperparameters, is based on PYMARL2 \cite{hu2023rethinkingtheimplementation} and CFCQL. Fine-tuning is conducted for 1 million environment steps in \textit{2s3z} and \textit{3s5z}, and for 3 million steps in \textit{5m\_vs\_6m} and \textit{6h\_vs\_8z}.

\textbf{Baselines.}We compare OVMSE against four baselines: (i) MACQL \cite{CQL}, which uses CQL loss throughout both the offline and online phases; (ii) MACal-QL, which applies Cal-QL (an algorithm designed for O2O single-agent RL) \cite{Cal-QL} to MARL; (iii) QMIX, trained from scratch \cite{QMIX}; and (iv) Switch CQL in \textit{3s5z}, \textit{5m\_vs\_6m}, and \textit{6h\_vs\_8z}, which switches off the CQL loss during the online phase. All baselines use QMIX as the backbone.

\textbf{Mixing Ratio and Hyperparameters.} Following \cite{Cal-QL}, we introduce the concept of mixing ratio, a value ranging from 0 to 1, representing the percentage of offline data used during online training. For instance, a mixing ratio of 0 means no offline data is used during online training, while a ratio of 0.5 means half the data comes from offline and half from online interactions. In this work, we select the optimal mixing ratio hyperparameters for all methods from the set $\{{0.0, 0.1, 0.3, 0.5}\}$. Our experiments show that OVMSE performs best with a mixing ratio of 0.0 or 0.1. The detailed hyperparameter settings can be found in Appendix.

\textbf{Results.} We report the median test win rates over environment steps in Figure~\ref{result figure} and the mean test return in Figure~\ref{test return}. All results are averaged across 6 random seeds. Additionally, we conduct 640 tests for each seed after online fine-tuning, with the final median test win rates presented in Table \ref{final performance}.
\begin{table*}[h!]
\caption{
The final median win rates for different tasks across algorithms, except for \textit{2s3z} in which all algorithms achieve near $100\%$ win rate. The results are obtained based on $640$ tests for each seed after the completion of online fine-tuning. } \label{final performance}
\centering
\begin{tabular}{|l|c|c|c|c|c|}
\hline
Task & OVMSE & Switch CQL & MACal-QL & MACQL & QMIX \\
\hline
6h\_vs\_8z medium replay & \textbf{0.71250} & 0.50859 & 0.11641 & 0.12969 & 0.32031 \\
6h\_vs\_8z medium  & \textbf{0.60078} & 0.53359 & 0.37187 & 0.34141 & 0.32031 \\
5m\_vs\_6m medium & \textbf{0.69531} & 0.63203 & 0.24688 & 0.24453 & 0.64063 \\
5m\_vs\_6m medium replay & 0.69297 & \textbf{0.71484} & 0.21641 & 0.20234 & 0.64063 \\
3s5z medium & \textbf{0.86719} & 0.74766 & 0.12969 & 0.17188 & 0.53359 \\
3s5z medium replay & \textbf{0.80625} & 0.74687 & 0.11797 & 0.04100 & 0.53359 \\
\hline
\end{tabular}

\end{table*}

\begin{figure*}[htbp]
    \centering
    \subfloat[]{\includegraphics[width=0.33\textwidth]{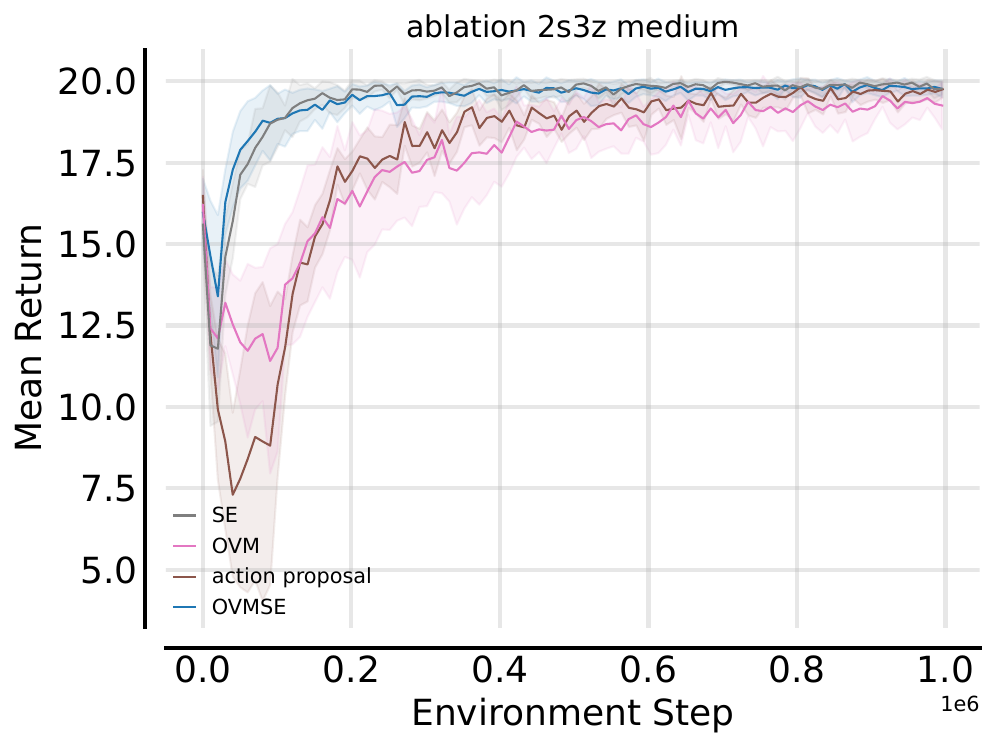}} 
    \subfloat[]{\includegraphics[width=0.33\textwidth]{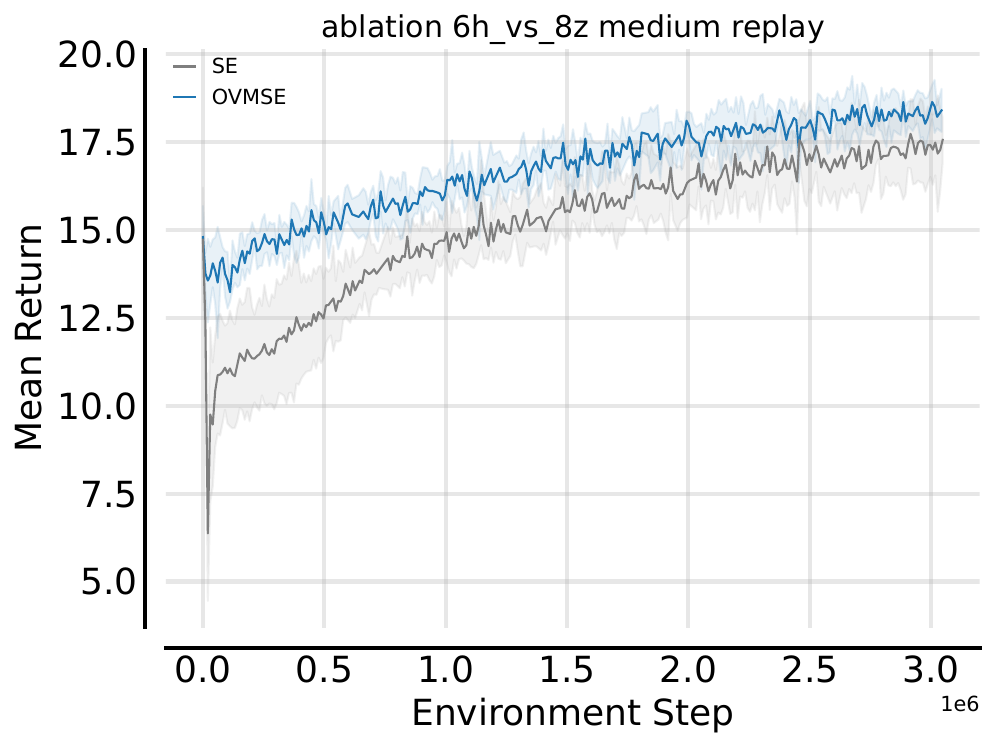}} 
    \subfloat[]{\includegraphics[width=0.33\textwidth]{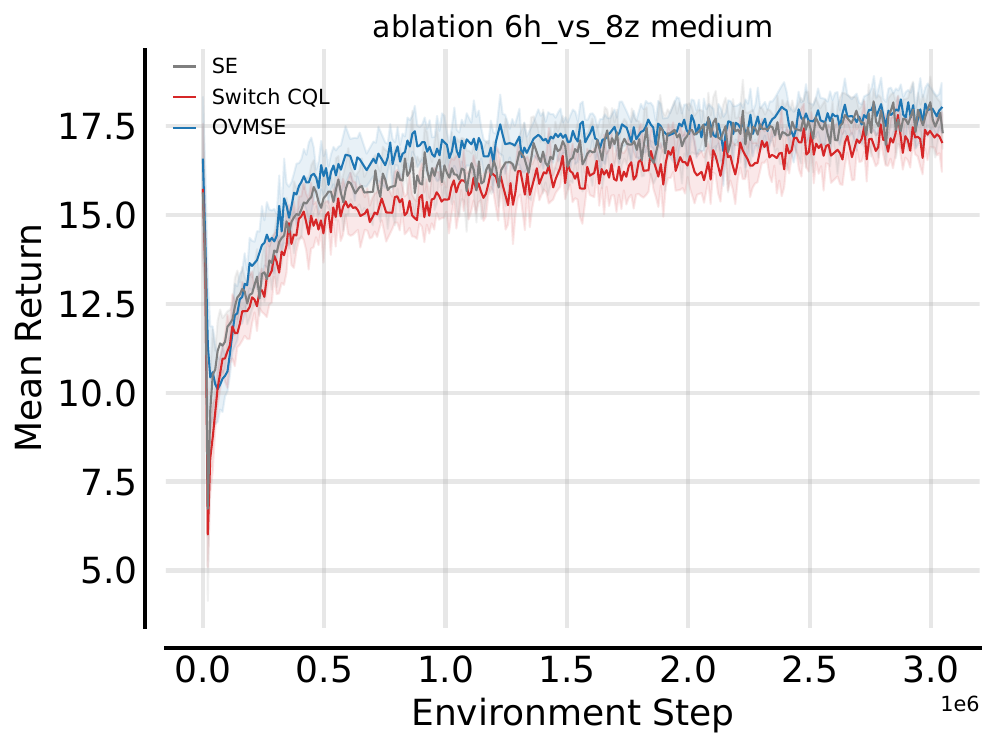}}
    \caption{We present the results of the ablation study and conclude that both OVM and SE are useful and effective.}
    \label{Ablation}
\end{figure*}

\begin{table*}[h!]

\centering
\begin{tabular}{|l|c|c|c|c|c|}
\hline
Task & Mixing Ratio 0.0  & Mixing Ratio 0.1  & Mixing Ratio 0.3 \\
\hline
2s3z medium  & 19.58 & 19.64 & 18.41 \\
\hline

\end{tabular}
\caption{
We conduct an ablation study for OVMSE with different mixing ratios on 2s3z medium and provide the mean test return at 200k environment steps. The results show that OVMSE has a minimal reliance on offline data (performing best with a mixing ratio of 0.0 and 0.1), as it preserves the offline learned Q-values.} \label{ablation on mixing ratio}
\end{table*}
\subsection{Results}

Figure \ref{result figure} and \ref{test return} present the training curves across four different tasks, pre-trained using both the medium and medium replay datasets. Our results show that OVMSE enables fast online fine-tuning and achieves superior overall performance. We compare OVMSE against baseline methods across three key aspects: (i) online performance, (ii) sample efficiency, and (iii) performance drop during the offline-to-online transition.

Table \ref{final performance} shows the final fine-tuned performance for all tasks (except for \textit{2s3z}, where all algorithms achieve near $100\%$ win rates). In the \textit{6h\_vs\_8z} medium replay task, OVMSE outperforms QMIX and Switch-CQL by more than $20\%$ in win rate. Similarly, in the \textit{3s5z} medium tasks, OVMSE leads other baselines by roughly $10\%$ in win rate, and in the \textit{3s5z} medium replay, \textit{6h\_vs\_8z} medium, and \textit{5m\_vs\_6m} medium tasks, OVMSE demonstrates a lead of more than $5\%$ in test win rate.

In terms of sample efficiency, OVMSE performs significantly better than the baselines. For instance, in the \textit{6h\_vs\_8z} medium and medium replay tasks, OVMSE achieves a $40\%$   win rate approximately $1.5$ million environment steps ahead of other baselines. In the \textit{5m\_vs\_6m} medium task, OVMSE achieves a $40\%$ win rate about $0.5$ million environment steps earlier than QMIX and Switch-CQL, with these baselines needing a total of $1.5$ million additional steps to match OVMSE’s performance. In the \textit{2s3z} medium and medium replay tasks, OVMSE achieves an $80\%$ win rate with a lead of more than $100k$ environment steps over MACQL, MACal-QL, and QMIX.

Furthermore, OVMSE experiences significantly less performance drop compared to all baselines during the offline-to-online transition. Notably, although other baselines typically use a higher mixing ratio than OVMSE, OVMSE still exhibits significantly less performance drop in the \textit{6h\_vs\_8z} medium, medium replay, \textit{2s3z} medium and medium replay, and \textit{5m\_vs\_6m} medium and medium replay tasks. This difference in performance drop is more evident in terms of mean test return, which is shown in Figure \ref{test return}. These results show that our OVMSE algorithm efficiently utilizes offline data and avoids unlearning. 

\subsection{Ablation Study}

In this section, we conduct additional experiments to demonstrate the effectiveness of both OVM and Sequential Exploration. Through ablation studies, we aim to answer three key questions:
\begin{enumerate}
     \item[(i)] Does OVM preserve offline knowledge and lead to better performance?
     \item[(ii)] Does SE enable efficient exploration and lead to better performance?
     \item[(iii)] Does OVMSE rely on reusing offline data during online phase?
\end{enumerate}

We perform these ablation studies on the \textit{2s3z} medium task, \textit{6h\_vs\_8z} medium and medium replay tasks. As shown in Figure~\ref{Ablation}(a), (b), and (c), we set the mixing ratio to 0.0 for all tasks except for the \textit{6h\_vs\_8z} medium task, where a mixing ratio of 0.1 is applied. Additionally, we conduct a specific ablation study on the \textit{2s3z} medium task to examine the effect of reusing offline data during the online phase. As illustrated in Figure~\ref{Ablation}, we report the mean test returns as a function of environment steps, as this metric more effectively captures performance drops during the initial stages of online learning.

\textbf{OVM is effective.} Our results demonstrate that OVM plays a crucial role in O2O MARL. In the \textit{2s3z} medium task, using SE alone without OVM results in a more significant performance drop during the initial stage of the online phase. Additionally, using OVM alone results in a much smaller performance drop compared to the action proposal approach \cite{ImitationBoostrap}. In the \textit{6h\_vs\_8z} medium replay task, OVMSE significantly outperforms SE alone, due to the preservation of offline knowledge by OVM. Similarly, in the \textit{6h\_vs\_8z} medium task, OVMSE outperforms SE, with SE requiring 2 million steps to catch up.

\textbf{SE is effective.} Our results also demonstrate that SE is highly effective. In the \textit{2s3z} medium task, OVMSE outperforms OVM alone by a large margin in the first $200k$ steps, due to SE’s more efficient exploration. In the \textit{6h\_vs\_8z} medium task, using SE alone outperforms Switch CQL, where the only difference between the two algorithms is the use of $\epsilon$-greedy exploration versus decentralized SE. This clearly demonstrates the effectiveness of SE.

\textbf{OVMSE has minimal reliance on offline data during online training.} We conduct an ablation study on 2s3z medium to examine the effect of reusing offline data during online training for OVMSE. We report the test returns at 200k environment steps (instead of the final fine-tuned test returns, as different mixing ratio values all achieve an optimal final test return). As shown in Table~\ref{ablation on mixing ratio}, OVMSE with a mixing ratio of 
0.0  and 0.1 achieves fast and stable online fine-tuning during the online phase, demonstrating that OVMSE is not sensitive to data distribution shift and effectively preserves offline learned Q-values. This highlights a clear advantage of OVMSE, as over-reliance on offline data can lead to slow online fine-tuning, especially when the offline data is of poor quality.


\subsection{Discussions}
In this section, we discuss the role of offline pre-trained knowledge in O2O MARL. Offline knowledge is beneficial in several ways. First, it helps guide online exploration much more efficiently than training from scratch. It also provides a strong initial policy. Furthermore, the offline pre-trained knowledge can facilitate faster online learning and lead to better performance, as clearly demonstrated in the \textit{6h\_vs\_8z} medium replay task shown in Figure~\ref{Ablation} (b), where OVMSE outperforms SE by a large margin, thanks to its offline knowledge. However, we also observe that offline knowledge can hinder online learning in some cases.  This issue likely arises because current offline training algorithms are not designed for the online phase. A good offline policy requires some degree of over-fitting to the offline datasets. However, for O2O MARL, this over-fitting can cause online learning to require many samples to correct or even result in the agent sticking to the offline policy, thereby preventing improvement. This observation motivated the introduction of the memory coefficient $\lambda_{\text{memory}}$ and its corresponding annealing schedule, which allow OVMSE to gracefully adjust offline knowledge to the level of online returns.


\section{CONCLUSIONS}
In this work, we investigate the O2O MARL. We identify two major challenges for O2O MARL: (i) During the initial stage of online learning, distributional shift can degrade the offline pre-trained Q-values, leading to unlearning of previously acquired knowledge; (ii) Inefficient exploration in the exponentially large joint state-action space can slow down online fine-tuning. To address these challenges, we introduce Offline Value Function Memory (OVM) to preserve offline knowledge and decentralized sequential exploration strategy for efficient online exploration. Our approach, OVMSE, enables stable and fast online fine-tuning and leads to improved fine-tuned results. Experiments across various tasks in SMAC demonstrate the superior performance of OVMSE.

\begin{acks}
This work is supported by the National Natural Science Foundation of China Grants 52450016 and 52494974. 
\end{acks}

\bibliographystyle{ACM-Reference-Format} 
\bibliography{sample}

\clearpage 
\onecolumn 

{\LARGE Supplementary Materials}

\appendix

\section{Detailed Hyperparameters}

In this section, we list in detail the hyperparameters used for OVMSE and all baseline algorithms.

\subsection{Shared Hyperparameters}  
For all methods, we aim to keep the majority of hyperparameters consistent with the PYMARL2 implementation \cite{hu2023rethinkingtheimplementation}, as these hyperparameters have been carefully fine-tuned. Table \ref{shared_hyperparameters} outlines the differences in hyperparameters compared to PYMARL2, along with the coefficients related to offline training. These hyperparameters are shared across all algorithms, with the exception of QMIX training from scratch. For QMIX, we retain all hyperparameters identical to those in PYMARL2, except for Batch\_Size\_Run and Target\_Update\_Interval, which we adjust to match the other baseline methods for a fair comparison.

\begin{table*}[h!]
\caption{Shared hyperparameter for different tasks.} \label{shared_hyperparameters}
\centering
\begin{tabular}{|l|cccc|}
\hline
\multirow{2}{*}{\textbf{Hyperparameters}} &\multicolumn{4}{c|}{\textbf{Tasks}}\\
                \cline{2-5}
                &\textit{2s3z}& \textit{3s5z} & \textit{5m\_vs\_6m}& \textit{6h\_vs\_8z} \\
\hline

Batch\_Size\_Run            & 2     & 2    & 8  & 8  \\
Target\_Update\_Interval    & 50   & 50  & 200 & 200 \\
Online\_Gradient\_Norm\_Clip & 1     & 0.5  & 10 & 10 \\
Offline\_CQL\_Weight        & 0.05  & 0.25 & 50 & 20 \\
Offline\_Training\_Steps (training steps)    & 3M    & 3M   & 10M & 10M \\
Online\_Training\_Steps (env steps)     & 1M    & 1M   & 3M  & 3M  \\
$\lambda_{\text{TD}}$       & 0.6   & 0.6  & 0.6 & 0.3 \\
\hline
\end{tabular}
\end{table*}
\subsection{Hyperparameters for Online Fine-tuning}
For online fine-tuning, we mainly adjusted the online CQL weight, the mixing ratio and the exploration coefficient.

\textbf{Mixing Ratio.}  
For all algorithms except QMIX, we perform a search on the mixing ratio from the set $\{0.0, 0.1, 0.3, 0.5\}$ and choose the best mixing ratio for each algorithm. The results are shown in Table \ref{Mixing Ratio}.

\textbf{Exploration Coefficient.}  
For all algorithms except QMIX, we use $\epsilon_{\text{start}}=1$ and $\epsilon_{\text{end}}=0.05$ in all O2O tasks. For QMIX, the hyperparameters for $\epsilon$-greedy exploration are the same for all tasks as in PYMARL2 \cite{hu2023rethinkingtheimplementation}.

\textbf{Online CQL Weight.} For MACQL and MACal-QL, we set the online CQL weight equal to the offline CQL weight for all tasks. OVMSE uses the same online CQL weight for \textit{2s3z} and the same online CQL weight as Switch CQL for the remaining tasks. The detailed settings are shown in Table \ref{online cql weight}.
\begin{table*}[h!] 
\caption{The mixing ratio of different algorithms for all datasets.}\label{Mixing Ratio}
\centering
\begin{tabular}{|l|c|c|c|c|}
\hline
\textbf{Datasets} & \textbf{OVMSE} & \textbf{Switch CQL} & \textbf{MACal-QL} & \textbf{MACQL}\\
\hline
\textit{2s3z}-m              & 0.0 & N/A & 0.5 & 0.3 \\
\textit{2s3z}-mr       & 0.0 & N/A & 0.3 & 0.3 \\
\textit{3s5z}-m              & 0.1 & 0.5 & 0.5 & 0.3 \\
\textit{3s5z}-mr       & 0.1 & 0.5 & 0.3 & 0.3 \\
\textit{5m\_vs\_6m}-m        & 0.1 & 0.1 & 0.5 & 0.5 \\
\textit{5m\_vs\_6m}-mr & 0.0 & 0.3 & 0.5 & 0.5 \\
\textit{6h\_vs\_8z}-m        & 0.1 & 0.1 & 0.5 & 0.5 \\
\textit{6h\_vs\_8z}-mr & 0.0 & 0.1 & 0.5 & 0.5 \\
\hline
\multicolumn{5}{l}{-m means medium, -mr means medium-replay.}
\end{tabular}
\end{table*}

\begin{table*}[h!]
\caption{The online CQL weight of different algorithms for all datasets.} \label{online cql weight}
\centering
\begin{tabular}{|l|c|c|c|c|}
\hline
\textbf{Datasets} & \textbf{OVMSE} & \textbf{Switch CQL} & \textbf{MACal-QL} & \textbf{MACQL}\\
\hline
\textit{2s3z}-m              & 0.05 & N/A & 0.05 & 0.05 \\
\textit{2s3z}-mr       & 0.05 & N/A & 0.05 & 0.05 \\
\textit{3s5z}-m              & 0.0 & 0.0 & 0.25 & 0.25 \\
\textit{3s5z}-mr       & 0.0 & 0.0 & 0.25 & 0.25 \\
\textit{5m\_vs\_6m}-m        & 0.0 & 0.0 & 50 & 50 \\
\textit{5m\_vs\_6m}-mr & 0.0 & 0.0 & 50 & 50 \\
\textit{6h\_vs\_8z}-m        & 0.0 & 0.0 & 20 & 20 \\
\textit{6h\_vs\_8z}-mr & 0.0 & 0.0 & 20 & 20 \\
\hline
\multicolumn{5}{l}{-m means medium, -mr means medium-replay.}
\end{tabular}
\end{table*}

\subsection{Specific Hyperparameters for OVMSE}
Featuring Offline Value Memory, the memory coefficient plays a crucial role in the performance of OVMSE. Table \ref{memory schedule} provides the detailed settings of $\lambda_{\text{memory}}$ and its annealing schedule across different datasets. Additionally, Table \ref{exploration params} lists the exploration hyperparameters used by OVMSE with Sequential Exploration (SE).

\begin{table*}[h!] 
\caption{The memory coefficient annealing schedule for all datasets.} \label{memory schedule}
\centering
\begin{tabular}{|l|c|c|c|}
\hline
\textbf{Datasets} & \textbf{$\lambda_{\text{memory\_start}}$} & \textbf{$\lambda_{\text{memory\_end}}$} & \textbf{Annealing Duration $T$}\\
\hline
\textit{2s3z}-m             & 1.0 & 1.0  & N/A  \\
\textit{2s3z}-mr            & 1.0 & 1.0  & N/A  \\
\textit{3s5z}-m             & 1.0 & 1.0  & N/A  \\
\textit{3s5z}-mr            & 1.0 & 1.0  & N/A  \\
\textit{5m\_vs\_6m}-m       & 1.0 & 0.1  & 200k \\
\textit{5m\_vs\_6m}-mr      & 1.0 & 0.05 & 200k \\
\textit{6h\_vs\_8z}-m       & 1.0 & 0.05 & 100k \\
\textit{6h\_vs\_8z}-mr      & 1.0 & 0.05 & 100k \\
\hline
\multicolumn{4}{l}{-m means medium, -mr means medium-replay.}
\end{tabular}
\end{table*}

\begin{table*}[h!]
\caption{Exploration hyperparameters for OVMSE with Sequential Exploration (SE).} \label{exploration params}
\centering
\begin{tabular}{|l|c|}
\hline
\textbf{Hyperparameters} & \textbf{Value} \\
\hline
Exploration Strategy     & Sequential Exploration (SE) \\
$\epsilon_{\text{start}}$   & 1.0 \\
$\epsilon_{\text{end}}$     & 0.05 \\
$\epsilon$ Annealing Duration & 100k timesteps \\
\hline
\end{tabular}
\end{table*}


\subsection{Offline CQL Algorithm Choice}  
The CQL-based offline MARL algorithm we choose for \textit{2s3z} and \textit{3s5z} is CQL, while we use CFCQL \cite{CFCQL} for \textit{5m\_vs\_6m} and \textit{6h\_vs\_8z} due to its better offline performance. All algorithms use the same CQL loss for the same tasks.

%

\section{Details on Data Collection}
\begin{figure}[H]
    \centering
    \subfloat[]{\includegraphics[width=0.35\textwidth]{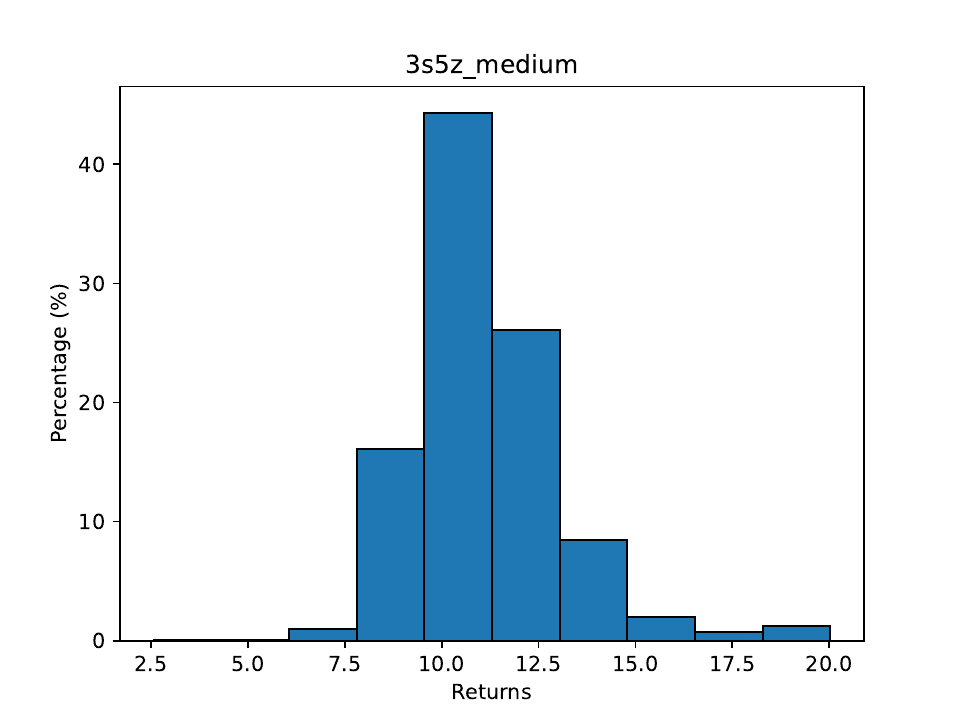}} 
    \subfloat[]{\includegraphics[width=0.35\textwidth]{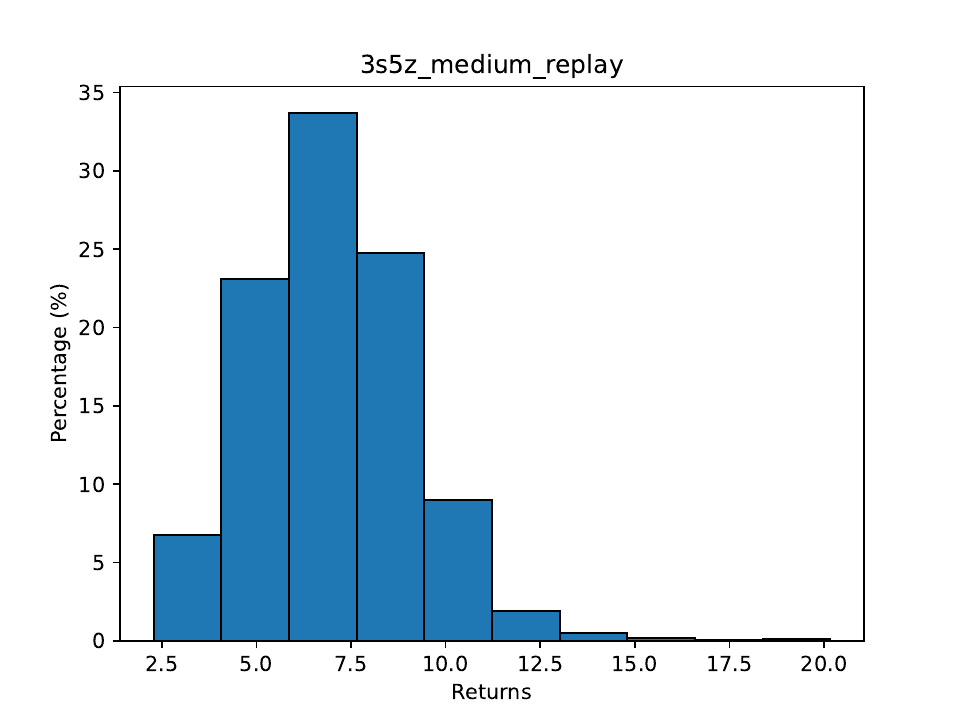}} 
    \quad
\caption{The distribution of trajectories with different rewards in the 3s5z medium and 3s5z medium-replay datasets. The horizontal axis represents the return of each trajectory, and the vertical axis represents the percentage of trajectories corresponding to that return.}
    \label{data_dist}
\end{figure}
Since an offline dataset for the SMAC \cite{grandmasterstarcraft2} \textit{3s5z} task is currently unavailable, we created our own medium and medium-replay datasets. First, we trained a multi-agent policy from scratch to a medium level by running QMIX, implemented by PYMARL2, for $100,000$ environment interaction steps. For the medium dataset, we executed the trained medium-level policy in the environment and collected $5,000$ episodes. For the medium-replay dataset, we saved the replay buffer from the QMIX agent at the final training step (i.e., the $100,000$th step). Details of the datasets are provided in Table \ref{data_info}, and the distribution of the datasets are shown in Figure \ref{data_dist}.
\begin{table*}[h!]
\caption{Primary information about our own dataset.} \label{data_info}
\centering
\begin{tabular}{|l|c|c|c|c|}
\hline
\textbf{Datasets} & \textbf{Size} &  \textbf{Min Return}&\textbf{Average Return} & \textbf{Max Return}\\
\hline
\textit{3s5z} medium  & 5,000 Episodes& 2.57 & 11.15 & 20.03 \\
\textit{3s5z} medium-replay       & 1,908 Episodes& 2.29 & 7.05 & 20.16 \\
\hline
\end{tabular}
\end{table*}

\subsection{Important Notes for Reproduction}
We highlight several hyperparameters that are critical for reproducing OVMSE's reported performance:

\textbf{Offline Training Budget.}  
For \textit{5m\_vs\_6m} and \textit{6h\_vs\_8z}, the offline training phase runs for \textbf{10M training steps}, consistent with the default configuration of the CFCQL codebase\footnote{\url{https://github.com/thu-rllab/CFCQL}} upon which our implementation is built. For \textit{2s3z} and \textit{3s5z}, we use 3M training steps. As a reference, the CFCQL paper \cite{CFCQL} reports offline win rates of \textbf{21\%} on \textit{6h\_vs\_8z} medium-replay and \textbf{41\%} on \textit{6h\_vs\_8z} medium with \textbf{10M} training steps; reducing the offline training budget to fewer steps can result in substantially lower offline model quality. Since the quality of the offline-trained value function is fundamental to the Offline Value Memory (OVM) mechanism ,an insufficiently trained offline model will produce unreliable value estimates, significantly degrading OVM's effectiveness and the initial policy quality that Sequential Exploration (SE) depends on. 

\textbf{Mixing Ratio.}  
As shown in Table \ref{Mixing Ratio}, OVMSE uses a mixing ratio of 0.0 for most datasets and 0.1 for the remainder. This is a deliberate design choice: OVM is intended to replace the reliance on offline data during online fine-tuning by providing value-level guidance instead of data-level replay. Overriding this setting with a higher mixing ratio may degrade performance.


\end{document}